\documentclass[letterpaper, 10 pt, journal, twoside]{IEEEtran}
\IEEEoverridecommandlockouts                              % This command is only needed if 
\newcommand{\blue}[1]{#1}

\newcommand{\vts}[1]{\lvert #1 \rvert}

\newcommand\inv[1]{#1\raisebox{1.05ex}{$\scriptscriptstyle-\!1$}}
\newcommand\Tstrut{\rule{0pt}{2.6ex}}         % = `top' strut
\newcommand\Bstrut{\rule[-1.3ex]{0pt}{0pt}}   % = `bottom' strut
\newcommand{\pc}{\zeta_c(t)}
\newcommand{\pd}{\zeta_d(t)}

\makeatletter
\newcommand\footnoteref[1]{\protected@xdef\@thefnmark{\ref{#1}}\@footnotemark}
\makeatother

\newcommand{\shorteq}{%
  \settowidth{\@tempdima}{-}% Width of hyphen
  \resizebox{\@tempdima}{\height}{=}%
}

\usepackage{amssymb,fge}

\usepackage{amsmath, amssymb}
\usepackage{subcaption}
\usepackage[utf8]{inputenc}
\usepackage[T1]{fontenc}
\usepackage[dvipsnames]{xcolor}
\usepackage[linesnumbered,ruled,vlined]{algorithm2e}
\usepackage[font=small]{caption} % This causes problems while compiling to latex    
\usepackage{array}
\usepackage{graphicx}
\usepackage{amsfonts}
\usepackage{soul}
\usepackage{hhline}
\usepackage{multirow, makecell}
\usepackage{float}
\usepackage{booktabs}
\usepackage{anyfontsize}

\usepackage{amsthm}
\usepackage{color}
\usepackage{transparent}
\usepackage{url}
\usepackage{footmisc}
\usepackage{setspace}
\usepackage{enumitem}
\usepackage{mathtools}

\DeclarePairedDelimiter\abs{\lvert}{\rvert}

\DeclareMathOperator*{\argmax}{arg\,max}

\theoremstyle{plain}

\DeclareMathOperator{\EX}{\mathbb{E}}% expected value
\usepackage{graphicx} % for pdf, bitmapped graphics files
\usepackage{amsmath} % assumes amsmath package installed
\usepackage{amssymb}  % assumes amsmath package installed
\usepackage[normalem]{ulem}
\usepackage[colorlinks, citecolor=magenta]{hyperref}
\title{B-GAP: Behavior-Rich Simulation and Navigation for Autonomous Driving}

\author{Angelos Mavrogiannis, Rohan Chandra, and Dinesh Manocha% <-this % stops a space
\thanks{Manuscript received: February, 2, 2022; Accepted February, 6, 2022.}%Use only for final RAL version
\thanks{This paper was recommended for publication by Editor Tamim Asfour upon evaluation of the Associate Editor and Reviewers' comments. 
This work was supported in part by ARO Grants W911NF1910069, W911NF2110026, U.S. Army Grant No. W911NF2120076  and, Semiconductor Research Corporation (SRC) and Intel.} %Use only for final RAL version
\thanks{All authors are with the Department of Computer Science, University of Maryland, College Park, MD, 20742, USA
        {\tt\footnotesize angelosm@cs.umd.edu}}%
\thanks{Digital Object Identifier (DOI): see top of this page.}
}
\begin{document}

\maketitle
% \thispagestyle{empty}
% \pagestyle{empty}
% Comment or remove these lines for final RAL version.

%%%%%%%%%%%%%%%%%%%%%%%%%%%%%%%%%%%%%%%%%%%%%%%%%%%%%%%%%%%%%%%%%%%%%%%%%%%%%%%%
\begin{abstract}
We address the problem of ego-vehicle navigation in dense simulated traffic environments populated by road agents with varying driver behaviors. Navigation in such environments is challenging due to unpredictability in agents' actions caused by their heterogeneous behaviors. We present a new simulation technique consisting of enriching existing traffic simulators with behavior-rich trajectories corresponding to varying levels of aggressiveness. We generate these trajectories with the help of a driver behavior modeling algorithm. We then use the enriched simulator to train a deep reinforcement learning (DRL) policy that consists of a set of high-level vehicle control commands and use this policy at test time to perform local navigation in dense traffic. Our policy implicitly models the interactions between traffic agents and computes safe trajectories for the ego-vehicle accounting for aggressive driver maneuvers such as overtaking, over-speeding, weaving, and sudden lane changes. Our enhanced behavior-rich simulator can be used for generating datasets that consist of trajectories corresponding to diverse driver behaviors and traffic densities, and our behavior-based navigation scheme can be combined with state-of-the-art navigation algorithms.
\end{abstract}

% Keywords appear just beneath the abstract. Use only for final RAL version.  
\begin{IEEEkeywords}
Autonomous Agents, Autonomous Vehicle Navigation, Behavior-Based Systems, Intelligent Transportation Systems, Reinforcement Learning
\end{IEEEkeywords}
\section{Introduction}
% Drop letter for first word of the Introduction 
% Here we have the typical use of a "T" for an initial drop letter
% and "HIS" in caps to complete the first word.
% Use only for final RAL versionFormatting
\IEEEPARstart{T}{he} navigation problem for autonomous vehicles (AVs) corresponds to computing the optimal actions that enable the AV to begin from starting point $A$ and reach destination $B$ via a smooth trajectory while avoiding collisions with dynamic obstacles or traffic agents. A key aspect of navigation is safety because the AVs are expected to keep a safe distance from other vehicles while also making driving more fuel- and time-efficient. Navigation is a central task in autonomous driving, and navigation problems have also been studied extensively in the contexts of motion planning and mobile robots.

There is considerable research on designing prediction and navigation algorithms for autonomous driving, but these algorithms are currently primarily deployed in specific driving scenarios~\cite{highway2} or low-density traffic~\cite{highway1}. Some of these algorithms are intentionally designed to excel in unique cases~\cite{Song_2020}, while others are inevitably limited because they are based on data-driven methods trained on selected datasets with specific driving conditions~\cite{Deo_2018}. Modern prediction and navigation algorithms must be able to handle various driving scenarios to comply with real-world situations. One of these scenarios is dense traffic, which is commonly observed in city centers or in the vicinity of frequent or popular destinations. There are many challenges in terms of handling dense traffic environments, including computing safe and collision-free trajectories, and modeling the interactions between the traffic-agents. Some key issues are related to evaluating the driving behaviors of human drivers and ensuring that the driving pattern of the AV is consistent with traffic norms.  It has been observed that  current AVs tend to drive hyper-cautiously or in ways that can frustrate other human drivers~\cite{dirtyTesla-gamma}, potentially leading to fender-benders.
%While many research and industry efforts have resulted in practical navigation algorithms in sparse highway traffic conditions, the feasibility of these methods remains questionable in more challenging scenarios such as highly dense urban roads. 
Moreover, it is important to handle the unpredictability or aggressive nature of human drivers. For example, human drivers may act irrationally and move in front of other vehicles by suddenly changing lanes or aggressively overtaking~\cite{chandra2021meteor}. In 2016, Google's self-driving car had a collision with an oncoming bus during a lane change~\cite{davies2016google}. In this case, the ego-vehicle assumed that the bus driver was going to yield; instead, the bus driver accelerated.  Overall, we need better prediction and navigation methods that can account for such behaviors and more diverse datasets enriched with these behaviors so that learning-based methods can produce results that are more applicable to real-world scenarios.

\begin{figure}
    \centering
    \includegraphics[width = \columnwidth]{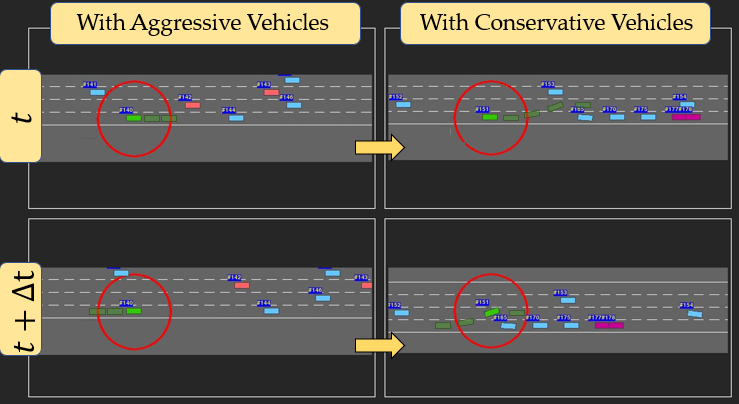}
    \caption{\blue{\textbf{Behavior-Rich Simulation and Navigation:}} We highlight the \blue{generated} actions and the trajectories 
    %trajectory (represented by \crule[ForestGreen]{.3cm}{.15cm}) rectangles)
    for the ego-vehicle (green) in aggressive (left) or conservative (right) environments. \textit{(Left)} At time $t$, the ego-vehicle identifies the aggressive vehicle (red) ahead in the left lane. \blue{Based on this information} the ego-vehicle \blue{anticipates} that, at time $t + \Delta t$, the aggressive vehicle \blue{might} make a lane change, and our \blue{learned policy} decides to slow down. \textit{(Right)} The ego-vehicle identifies a group of  conservative vehicles (cyan) in its path. Based on this conservative behavior, our \blue{learned policy} decides to change lanes by speeding up and overtaking the group of vehicles at time $t + \Delta t$. Overall, our  approach results in fewer collisions and improved navigation in complex scenarios \blue{compared to a default policy that does not account for different driver behaviors}.}
    \label{fig: cover}
    \vspace{-10pt}
\end{figure}

There is considerable work on classifying driver behaviors or styles in traffic psychology~\cite{quartz-article, schwarting2019social}. In most cases, driving style is defined with respect to either aggressiveness~\cite{doshi2010examining} or fuel consumption~\cite{corti2013quantitative}. Many recent methods have been proposed to classify driving behaviors based on past trajectories~\cite{rohanref3,ernestref10,cmetric} and use them for predicting the vehicle's future trajectory~\cite{chandra2020forecasting,li2019learning}. On the other hand, recent techniques for prediction and navigation have been based on reinforcement learning, and they tend to learn optimal policies with suitable rewards for collision handling, lane changing, and traffic rule adherence. Our goal is to extend these learning methods to account for driver behaviors.
%and generate 

%Current navigation methods have started using reinforcement learning in simulated environments to train agents to drive autonomously in traffic environments. However, these methods have mostly been tested in homogeneous scenarios that consist of conservative agents. To train an AV to navigate among aggressive human drivers, we must first introduce aggressive agents in a simulated environment. Driver behavior, however, is a complex and abstract notion that does not lend itself to a formal definition. Therefore, simulating aggressive agents is not a straightforward task of heuristically tuning the simulation environment parameters in a trial-and-error fashion, as is often the case with prior simulators.

%A recent line of research~\cite{cmetric} has proposed new techniques that combine graph theory, machine learning, and traffic psychology to model human driver behavior using raw trajectory data. The underlying approach consists of predicting the likelihood that a human driver performs certain maneuvers such as overspeeding, overtaking, sudden lane-changes, and weaving in realtime. While these techniques have been evaluated for their performance in driver behavior prediction, the possibility of using the underlying ideas for navigation has not been investigated.

\textbf{Main Contributions:} We present a new simulation technique of enriching traffic simulators with behavior-rich vehicle trajectories using driver behavior modeling methods~\cite{cmetric}. Once enriched, these simulators can generate lateral and longitudinal driver behaviors demonstrating varying levels of aggressiveness; for example, overspeeding, overtaking, lane-changing, and zig-zagging. Using this behavior-enriched simulator, we train a behaviorally-guided policy using deep reinforcement learning (DRL) \blue{that maps a state to a high-level vehicle control command}. Our \blue{control} policy considers the conservative or aggressive behavior of traffic agents \blue{and is paired with the simulator and its underlying controls and dynamics to perform local navigation}.
% We use a Markov Decision Process (MDP) to formulate the prediction of a high-level action for an AV (e.g., making a left or right lane change) in traffic environments consisting of aggressive drivers, and train a behavior-rich policy for safe ego-navigation.
%We also present techniques to model the reward distribution that is used to learn a behavior-rich policy that handles collision handling, lane-changing, and behavior-based traffic norms. 
Overall, the novel components of our work include:
\begin{enumerate}
    \item We use the CMetric algorithm~\cite{cmetric} to enrich existing traffic simulators with behavior-rich trajectories to generate realistic traffic simulations consisting of aggressive and conservative agents. Our technique is general and can generate traffic scenarios corresponding to varying traffic densities and driving styles.
    \item We use our behavior-rich simulator to train a DRL-based policy that implicitly models the interactions between agents and leads to improved navigation in dense traffic. Our formulation can automatically model the behavioral interactions between aggressive and conservative traffic agents and the ego-vehicle.
    \end{enumerate}
    We have integrated our enhanced simulation approach and policy training scheme into an OpenAI gym-based highway simulator~\cite{highway-env} \blue{paired with the Deep Q Learning algorithm~\cite{mnih2015human}} and have evaluated its performance in different kinds of traffic scenarios by varying the traffic density and the behaviors of traffic agents. However, our approach can be easily transferred to other state-of-the-art parametric traffic simulators~\cite{SUMO2018, carla} \blue{and combined with different learning algorithms.} We observe that our approach based on the behavior-guided simulator exhibits improved performance in dense traffic scenarios over the default OpenAI gym-based simulator which uses uniform driver behaviors. Our new simulation technique can serve as a diverse synthetic data-generating method for producing behavior-aware trajectories with a wide range of behaviors and varying traffic densities. Full set of results including comparison with state-of-the-art DRL methods can be found in the appendix at \href{https://arxiv.org/pdf/2011.03748.pdf}{\textbf{https://arxiv.org/pdf/2011.03748.pdf}}.
\section{Related Work}

\begin{figure*}
    \centering
    \includegraphics[width=.8\linewidth]{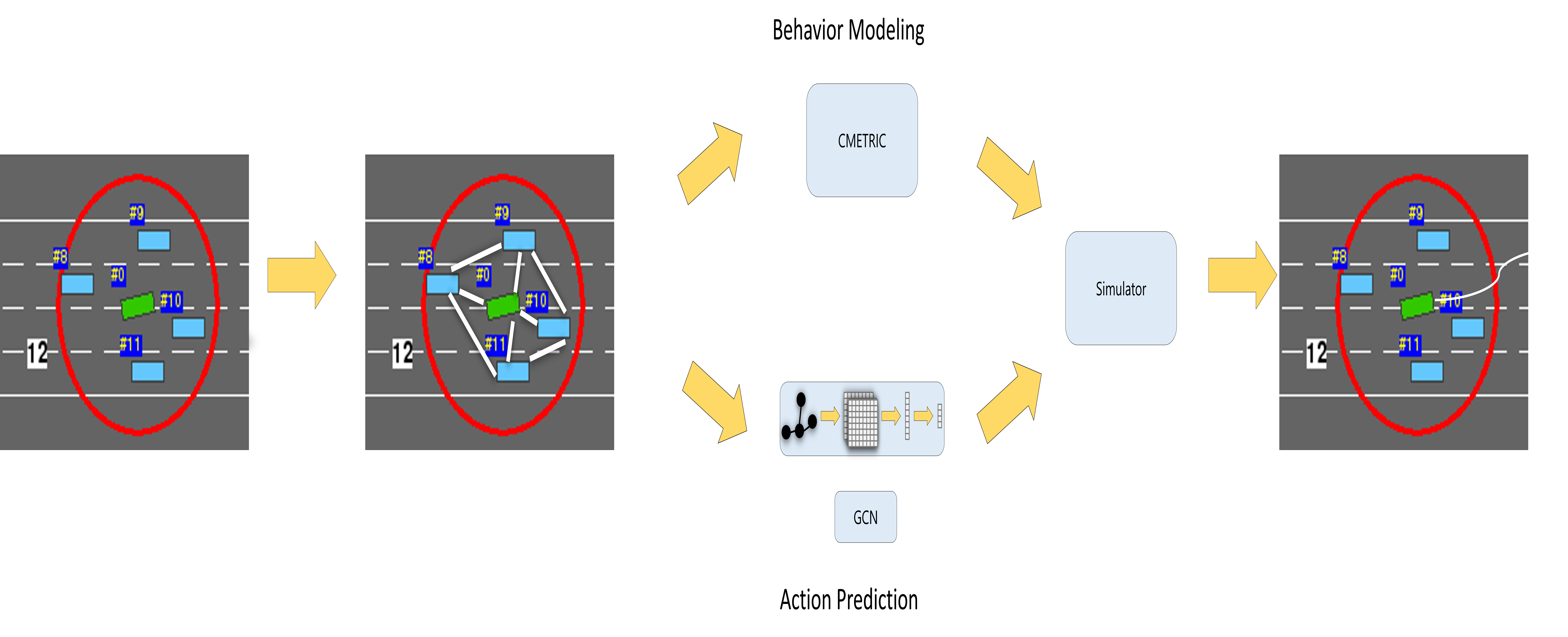}
    \caption{\textbf{Offline Training:} We highlight our new behavior-rich simulation technique and use it to generate a behavior-guided navigation policy for autonomous driving.
    In \textit{Step 1}, we use the CMetric behavior modeling algorithm to compute a set of parameters that characterize aggressive behaviors such as over-speeding, overtaking, and sudden lane changing. In \textit{Step 2},  we use these parameters to train a behavior-based navigation policy.} 
    \label{fig: offline_training}
    \vspace{-10pt}
\end{figure*}

\subsection{Navigation Algorithms}
\label{subsec: Related_Work:Navigation_Algorithms}
There is extensive work on navigation and planning algorithms in robotics and autonomous driving. At a broad level, they can be classified as techniques for
vehicle control~\cite{de1998feedback}, techniques for motion planning, and end-to-end learning-based methods. Vehicular control methods rely on a very accurate model of the vehicle, which needs to be known a priori for navigating at high speeds or during complex maneuvers. Motion planning methods are either algebraic or probabilistic search-based~\cite{lavalle2001randomized} or use non-linear control optimization~\cite{liniger2015optimization}. 
%However, handling complex obstacles and modeling the interactions between different road-agents remains a %challenge. 
%On the other hand, end-to-end motion planning methods that use supervised machine learning have been shown to be useful in eliminating many of the limitations of traditional methods. These learning-based methods, however, rely on large amounts of data in order to generalize well to new scenarios. We refer the reader to~\cite{schwarting2018planning} for a detailed review on learning-based and traditional methods.
Recently, many learning-based techniques have been proposed~\cite{corr2020,everett2017socially, cri-alahi}.
%To overcome the disadvantages mentioned above, recent interest has shifted to navigation using reinforcement learning
 These include reinforcement learning algorithms that aim to find an optimal policy that directly maps the sensor measurements to control commands such as velocity or acceleration and steering angle. Optimal policies for navigation can be learned with suitable rewards for collision handling, lane changing, and traffic rule and regulation adherence. However, the learned policies may not consider the behavior of other road entities or traffic entities.
 %aggressive driver behavior. 
 Some methods~\cite{everett2017socially} use a linear constant velocity model that assumes that road-agents move mostly in straight lines with fixed velocities. %for large time-steps.
 Other methods use deep learning-based trajectory prediction models to predict the future state~\cite{cri-alahi, chandra2019traphic, chandra2019robusttp}. However, the noisy inputs obtained from this model degrade the accuracy of the navigation method~\cite{goodfellow2014explaining}. Our behavior-based formulation can be integrated with most of these methods.%Other methods use collision avoidance policies based on reciprocal velocities and also give theoretical guarantees~\cite{van2011reciprocal}, but these are mainly limited to homogeneous road-agents represented as circular shapes. 

\subsection{Driver Behavior Modeling}
\label{subsec: Related_Work:Behavior_Modeling}

%At an high level, the literature on driver behavior studies can be characterized into three classifications. The principal classification of research arranges driver behavior dependent on the attributes of drivers, for example, age, sexual orientation, pulse, character, occupation, hearing, and so on. 
There is considerable work in traffic literature on classifying driver behavior based on driver attributes~\cite{rohanref5,rohanref3,ernestref10}.
Other sets of classification methods depend on natural factors, including climate or traffic conditions~\cite{behaviorref-category2-1} and the mental health of the drivers~\cite{behaviorref-category3-3}. Sagberg et al.~\cite{sagberg2015review} define driver behavior based on its underlying motives and separate it into global driving styles and specific driving styles. As an example of this taxonomy, drivers who are motivated to get to their destination as fast as possible exhibit certain driving styles such as overspeeding, honking, or tailgating. Based on these directly observable specific driving styles, these drivers are classified into the global driving style of aggressive driving. Although researchers have considered various driver behavior modeling techniques and thereby different numbers of categories of distinct driver behaviors, we ultimately adopt the simple and common distinction suggested by Sagberg et al., which separates driver behavior to aggressive/risky and conservative/defensive.
%The second classification of literature depends on natural factors, for example, climate or traffic conditions
%~\cite{behaviorref-category2-1,behaviorref-category2-2}. The examination conducted in~\cite{behaviorref-category2-2} was intended to research the impacts of climate-controlled speed-limits and traffic-signs on slippery roads. Other studies~\cite{behaviorref-category2-1} have associated changes in rush hour traffic with shifting driver behavior. 

%The last classification of examination alludes to mental health problems that influence driving styles. Mental health incorporate alcoholic driving, driving impaired, and condition of exhaustion. It is shown~\cite{behaviorref-category3-2} that driving impaired initiates postponed reactions in increasing speed and deceleration. Jackson et al.~\cite{behaviorref-category3-3} show that a condition of weariness shows similar attributes as driving impaired, however without the impact of substance inebriation. Different methods assess the effect of cell phone procedure on driver behaviors~\cite{behaviorref-category3-1}.

Many approaches have also been proposed for modeling driver behavior~\cite{schwarting2018planning}.
%In addition to these studies, numerous realtime approaches have also been proposed for driver behavior modeling. 
These strategies are based on partially observable Markov decision processes (POMDPs)~\cite{li2017game, cao2020reinforcement}, data mining techniques~\cite{constantinescu2010driving}, game theory~\cite{pnas, pnas1}, and imitation learning~\cite{kuefler2017imitating}. The approach we use~\cite{cmetric} is based on graph theory~\cite{chandra2021using, chandra2019graphrqi, chandra2020stylepredict} and only uses directly observable features to classify driver behaviors without making any assumptions about the psychological conditions and personalities of the drivers.
%While our action prediction algorithm uses the CMetric algorithm~\cite{cmetric}, it can also be combined with other driver behavior modeling methods.
%Interested readers may read Schwarting et al.~\cite{schwarting2018planning} for a more detailed survey of these strategies. The primary limitation of these strategies is that the motivations of human drivers are modeled probabilistically, thereby leading to low robustness. Probabilistic, or non-deterministic, approaches moreover depend on reasonable prior distributions that may not be accessible.

\subsection{\blue{Traffic Simulators}}
\blue{Traffic simulators are being widely used in autonomous driving research for testing and benchmarking new approaches and technologies. Many of these simulators model traffic from a bird's eye view~\cite{SUMO2018, highway-env, fluids}, while others adopt a more realistic approach that is closer to a human driver's point of view~\cite{carla} or data-driven methods~\cite{li2019aads}. While these simulators provide multiple options in terms of driving scenarios (highway, roundabout, intersection, etc.) and other advanced capabilities, they do not offer built-in functionality for generating diverse driving behaviors. Some of them~\cite{carla,highway-env} explicitly define aggressive and defensive agents, but their underlying policies are simple and rigid since they often rely on heuristic-based models that directly encode traffic rules. There have been some recent advances in generating behavior-rich driving policies~\cite{trafficsim,shiroshita2020behaviorally} using learning-based methods, but these methods broadly focus on generating a diverse set of driving behaviors without explicitly characterizing the generated trajectories. Our proposed simulation technique is different and explicitly considers aggressive and conservative behaviors. It does not require manual parameter adjustment and can be easily integrated into any of the state-of-the-art parametric traffic simulators.}
\section{ Problem Formulation and Background}
\label{sec: Problem_Formulate}
\subsection{Problem Formulation}
Our goal is \blue{to design a behavior-rich simulator and use our simulator} to train a behavior-rich policy for safe \blue{ego-navigation} in dense traffic. Such a behavior-rich navigation policy is trained using deep reinforcement learning.
% Given the current state of the environment, which may consist of the positions and velocities of all the traffic-agents at current time $t$, our goal is to predict the \blue{high-level} action for the ego-vehicle at the next time step $t+1$ using a trained policy. To this end, our prediction module is based on a Markov Decision Process (MDP), which characterizes a policy $\pi(u|s)$ that outputs the optimal action $a$ for an autonomous vehicle given the state of the environment $s$. This action typically belongs to a finite set of longitudinal (speed up, slow down, maintain speed) and lateral (turn left/right) actions.

\subsection{Markov Decision Processes (MDP) and Deep Q Networks}
A standard model used in deep reinforcement learning is the Markov Decision Process (MDP). 
% Autonomous vehicles must be able to predict high-level actions such as ``left'' or ``right'' to make safe and efficient decisions in real-time traffic.
An MDP $\mathcal{M}$ consists of states $s \in \mathcal{S}$, controls $u \in \mathcal{U}$, a reward function $\mathcal{R}(s, u)$, and a state transition matrix $\mathcal{T}(s^\prime|s, u)$ that computes the probability of the next state given the current state and control. A policy $\pi(u|s)$ represents a distribution for each state. The goal in an MDP is to find a policy that obtains high future rewards. For each state $s \in \mathcal{S}$, the agent executes a control $u \in \mathcal{U}$. Upon execution, the agent receives a reward $\mathcal{R}(s, u)$
and reaches a new state $s^\prime$, determined from the transition matrix $\mathcal{T}(s^\prime|s, u)$. A policy $\pi(u|s)$ specifies the control that the agent will execute in each state. The goal of the agent is to
find the policy $\pi(u|s)$ that maps states to controls to maximize
the expected discounted total reward over the agent’s lifetime. The value $Q^{\pi}(s, u)$ of a given state-control pair $(s, u)$
is an estimate of the expected future reward that can be obtained from $(s, u)$ when following policy $\pi$. The optimal
value function $Q^*(s, u)$ provides maximal values in all states
and is determined by solving the Bellman equation:

\[Q^*(s, u) = \EX \Bigg [ R(s, u) + \gamma \sum_{s^\prime}P(s^\prime | s, u) \max_{a^\prime}Q^*(s^\prime, u^\prime) \Bigg] \]

The optimal policy $\pi$ is then $\pi(s) = \argmax_{a \in \mathcal{U}} Q^*(s, u)$.
Deep Q Networks (DQNs)~\cite{mnih2015human} approximate the value function $Q(s, u)$ with a deep neural network that outputs a set of control values $Q(s,\boldsymbol{\cdot}; \theta)$ for a given state input $s$, where $\theta$ are the
parameters of the network.

% \subsection{MDP For Navigation in Aggressive Environments}
% \label{mdp}

\subsection{CMetric - A Driver Behavior Model}
To generate behavior-rich trajectories, we use CMetric~\cite{cmetric}, which is a parametric driver behavior classification algorithm. In the CMetric measure, the nearby traffic entities are represented using a {\em proximity graph}. The vertices denote the positions of the vehicles, and the edges denote the Euclidean distances between the vehicles in the global coordinate frame. Using centrality functions~\cite{rodrigues2019network}, CMetric presents techniques to model both longitudinal (along the axis of the road) and lateral (perpendicular to the axis of the road) aggressive driving behaviors.
%to a near human-level degree of accuracy. 

\subsubsection{Lateral Behaviors}
Sudden lane changing, overtaking, and weaving are some of the most typical lateral maneuvers made by aggressive vehicles. We model these maneuvers using the closeness centrality, which measures how centrally placed a given vertex is within the graph. The closeness centrality is computed by calculating the reciprocal of the sum of shortest paths from the given vertex to all other vertices.
% \begin{definition}
% \textbf{Closeness Centrality:} In a connected traffic-graph at time $t$ with adjacency matrix $A_t$, let $\mc{D}_t(v_i,v_j)$ denote the minimum total edge cost to travel from vertex $i$ to vertex $j$, then the discrete closeness centrality measure for the $i^\textrm{th}$ vehicle at time $t$ is defined as
% \begin{equation}
%     \zeta^i_c[t] = \frac{N-1}{\sum_{v_j\in \mathcal{V}(t)\setminus \{v_i\}} \mc{D}_t(v_i,v_j)},
%     \label{eq: closeness}
% \end{equation}
% % where $v \in \mathcal{V}$ denote all vertices in the connected traffic-graph other than $u$.
% \end{definition}
If a vertex is located more centrally, it results in a smaller sum, and thereby a higher value of the closeness centrality. The centrality for a given vehicle thus increases as the vehicle moves towards the center and decreases as it moves away from the center. Lateral driving styles (i.e., styles executed perpendicular to the axis of the road) are modeled using the closeness centrality.

% \begin{figure}
%     \centering
%     \includegraphics[width=\columnwidth]{Images/offline_training_mlp.PNG}
%     \caption{\textbf{Offline Training:} We highlight our behavior-guided navigation policy for autonomous driving. We use a behavior-rich simulator that can generate aggressive or conservative driving styles.
%     In \textit{Step 1}, we use the CMetric behavior classification algorithm to compute a set of parameters that characterize aggressive behaviors such as over-speeding, overtaking, and sudden lane changing. In \textit{Step 2},  we use these parameters to train a behavior-based action class navigation policy for action prediction and local navigation.} 
%     \label{fig: offline_training}
%     \vspace{-10pt}
% \end{figure}

\subsubsection{Longitudinal Behaviors}
Longitudinal aggressive driving behavior consists of over-speeding. We model the over-speeding by using the degree centrality, which measures the number of neighbors of a given vertex.
% \begin{definition}
% \textbf{Degree Centrality:} In a connected traffic-graph at time $t$ with adjacency matrix $A_t$, let $\mc{N}_i(t) = \{ v_j \in \mc{V}(t), \ A_t(i,j) \neq 0, \nu_j \leq \nu_i\}$ denote the set of vehicles in the neighborhood of the $i^\textrm{th}$ vehicle with radius $\mu$, then the discrete degree centrality function of the $i^\textrm{th}$ vehicle at time $t$ is defined as,
% \begin{equation}
%     % \zeta_d(u) = \vts{ \{ A(u,v)\in A_{u,:} \land  \nu_u \geq \nu_v \} },
%     \begin{aligned}
%     \zeta^i_d[t] = \bigl | \{ v_j \in \mc{N}_i(t) \} \bigr | + \zeta^i_d[t-1] &\\
%     \textrm{such that} \ (v_i,v_j) \not\in \mc{E(\tau)}, \tau = 0, \ldots, t-1&
%     \end{aligned}
%     % \zeta_d(u) = 
%     \label{eq: degree}
% \end{equation}
% where $\vts{\cdot}$ denotes the cardinality of a set and $\nu_i, \nu_j$ denote the velocities of the $i^\textrm{th}$ and $j^\textrm{th}$ vehicles, respectively. 
% \end{definition}
Intuitively, an aggressive or over-speeding vehicle will observe new vehicles (increasing degree) at a higher rate than a neutral or conservative vehicle. We use this property to distinguish between over-speeding and non-over-speeding vehicles.
\section{Offline Training}
\label{sec: Offline_Training}
In this section, we present our simulation technique for generating behavior-rich trajectories and our scheme for training a behavior-based navigation policy \blue{using this simulation technique}.
In order to learn such a policy, it is necessary to use a simulation environment with traffic agents that have varying behaviors, ranging from conservative to aggressive. These behaviors can be controlled using different parameters within a simulator, which are computed offline using a state-of-the-art driver behavior model~\cite{cmetric}. Our behavior-enriched simulator is used within a deep reinforcement learning (DRL) paradigm to learn a policy for behavior-guided navigation. We highlight this offline training setup in Figure~\ref{fig: offline_training}.
%In  the remainder of this Section, we present the details of our offline training setup.

\subsection{Enriching the Simulator with Driver Behaviors}
\label{subsec: Generating_Aggressive_Behaviors}
Many techniques have been proposed to classify driver behaviors~\cite{schwarting2018planning}. In this work, we use the CMetric~\cite{cmetric} measure to generate aggressive driver behaviors in a simulated environment. A vehicle's behavior in a simulator can be controlled through specific parameters. Depending on the values of these parameters, a vehicle's trajectory can exhibit aggressive or conservative behavior. In our parametric simulator, the problem of generating aggressive behaviors reduces to finding the appropriate range of parameter values that can generate realistically aggressive behaviors. The overall method for generating these parameters is summarized as follows:

\begin{enumerate}
    \item \textbf{Step 1 (Initialization):} At the first iteration, we begin by randomly choosing $\Xi(A)$, which represents the set of simulation parameters for generating aggressive agents. \label{algo: Algorithm}
    
    \item \textbf{Step 2 (Simulate behaviors):} $\Xi(A)$ is then passed as an input to the simulator to generate the trajectories for aggressive vehicles. These vehicles perform both longitudinal and lateral aggressive maneuvers. We compute the likelihood of these trajectories being aggressive using CMetric.
    
    \item \textbf{Step 3 (Feedback update):} Based on this likelihood feedback, we update the parameters, $\Xi(A)$, for the next iteration.
    
    \item We repeat this process until convergence.
\end{enumerate}

The final values from this procedure (Table~\ref{tab: parameters}) are used to generate aggressive and conservative trajectories. \blue{In the remainder of this section, we provide further details about our algorithm.}

\blue{In the first step, we initialize the parameters of the Highway-env simulator~\cite{leurent2019social} (described in Table~\ref{tab: parameters}) with random values. Our goal is to iteratively update these parameters so that the resulting trajectories that are generated by the simulator describe two types of drivers--aggressive and conservative. During each iteration, we first generate trajectories using the current parameter values and determine the behavior of these trajectories using the CMetric algorithm. The CMetric algorithm models the first- and second-order derivatives of these trajectories to output a final score that represents the level of aggressiveness of the trajectory. Further details can be found in Appendix I in the full version of the paper~\cite{mavrogiannis2021bgap}. For both the aggressive and conservative behavior class, we have a corresponding target CMetric value that each trajectory must reach in order to be categorized in that class. These targets are chosen heuristically. After generating a trajectory, we compare its CMetric score with the target score and update the underlying parameters accordingly. For example, for the aggressive class of behaviors, if the CMetric score for a particular trajectory is lower than its target score, then we would decrease the politeness factor and increase acceleration to increase the aggressiveness of the trajectory for the next iteration. This iterative process terminates once the trajectory reaches its target CMetric score.}

\subsection{Training a Behavior-Rich Policy}
\label{subsec: Training A Behavior-Rich Policy}
\begin{table}[t]
\caption{We use CMetric to obtain the simulation parameters that define the conservative and aggressive vehicle classes.}
\centering
\resizebox{\columnwidth}{!}{%
\begin{tabular}{cccc} 
\toprule
Model & Parameter \Tstrut & Conservative \Bstrut &   Aggressive \\
\hline
\multirow{4}{*}{IDM~\cite{treiber2000congested}}& Time gap ( $T$) \Tstrut & 1.5s      & 1.2s \\
 &Min distance ($s_0$) & 5.0 $m$ & 2.5 $m$ \\
&Max comfort acc. ($a$)     & 3.0 $m/s^2$ & 6.0 $m/s^2$\\
&Max comfort dec. ($b$) & 6.0 $m/s^2$               &  9.0 $m/s^2$ \\
\midrule
\multirow{3}{*}{MOBIL~\cite{kesting2007general}}& Politeness ($p$) & 0.5     & 0\\
& Min acc gain ($\Delta a_{th}$) & 0.2 $m/s^2$ & 0 $m/s^2$ \\
& Safe acc limit ($b_{safe}$) & 3.0 $m/s^2$ & 9.0 $m/s^2$\\
\bottomrule
\end{tabular}
}
\label{tab: parameters}
\vspace{-10pt}
\end{table}

We frame the underlying problem of learning a navigation policy for the ego-vehicle as a Markov Decision Process (MDP) represented by $\mathcal{M} := \{\mathcal{S},\mathcal{U},\mathcal{T},\gamma,\mathcal{R}\}$. The sets of possible states and controls are denoted by $\mathcal{S}$ and $\mathcal{U}$, respectively. $\mathcal{T}: \mathcal{S} \times \mathcal{S} \times \mathcal{U} \rightarrow \mathbb{R}$ captures the state transition dynamics,  $\gamma$ is the discounting factor, and $\mathcal{R}$ is the set of rewards defined for all of the states in the environment. We explain each of these further below:

\textbf{State Space:} The state of the world $S$ at any time step is equal to a matrix $F\times V$, which includes the state $s$ of every vehicle in the environment, where $s\in\mathbb{R}^F$. $V$ is the number of vehicles considered and $F$ is the number of features used to represent the state of a vehicle. \blue{$V$ is fixed for the entire duration of a set of experiments ($V=5$ for the sparse traffic case and $V=40$ for the dense case.)} $F$ includes the following:

\begin{itemize}
    \item \textit{presence}: \blue{a binary variable denoting whether a vehicle is observable at the current time step. A vehicle is considered observable if it lies within a specified distance from the ego-vehicle.}
    \item $x$: the longitudinal coordinate of the center of a vehicle.
    \item $y$: the lateral coordinate of the center of a vehicle.
    \item $v_x$: the longitudinal velocity of a vehicle.
    \item $v_y$: the lateral velocity of a vehicle.
\end{itemize}
\blue{In our experiments we consider a fully observable setting, but we also model a partially observable state where vehicles are only observable if they are within a smaller radius around the ego-vehicle, assuming limited ego-sensing capabilities.}

\textbf{Environment:} The environment consists of a highway with four \blue{unidirectional} lanes containing conservative and aggressive traffic agents generated using CMetric.

\textbf{Control Space:}  The ego-vehicle can execute five different controls: $\mathcal{U}=$\{"accelerate", "decelerate", "right lane-change", "left lane-change", "idle"\}.

\textbf{Reward Distribution:} The philosophy for orchestrating the reward function $\mathcal{R}$ is based on our objective for training an agent that can safely and efficiently navigate in dense traffic while respecting other road agents in its neighborhood. To this end, we use four rules to determine the reward of a state. At every time step $t$, collision with any (conservative or aggressive) road-agent earns the ego-vehicle $r_{C}$ points, lane changing earns $r_{LC}$ points, staying in the rightmost lane gives $r_{RL}$ points, and maintaining high speed increases $r_{HS}$:
\begin{equation*}
    \mathcal{R}^t=r_C^t+r_{LC}^t+r_{RL}^t+r_{HS}^t.
\end{equation*}
% Essentially, the key task during training was carefully tuning the reward values so that the ego agent can use lane changing wisely in order to avoid vehicles in its lane while also maintaining a safe and functional speed to ensure traffic flow is not disrupted.

% DO YOU EVALUATE THIS REWARD FUNCTION. CAN YOU PERFORM AN ABLATION STUDY, WHERE YOU REMOVE OR CONSIDER ONLY 2 OR 3 TERMS AND SEE THE IMPACT ON THE RESULTING RL METHOD PERFORMANCE

\textbf{State Transition Probability:} The state transition matrix $\mathcal{T}$ boils down to a state transition probability $P(s^\prime|s)$, which is defined as the probability of beginning from a current state $s$ and arriving at a new state $s^\prime$. This probability is calculated by the kinematics of the simulator, which depend on the underlying motion models (described further in Section~\ref{sec: Navigation_at_Runtime}), and thus it is equal to $1$, establishing a deterministic setting. 

% This means that if a road agent decides to take an action, e.g., if it attempts a lane change towards the left lane, it will execute the lane change successfully with a probability equal to $1$.

\begin{figure}[t]
    \centering
    \includegraphics[width = \columnwidth]{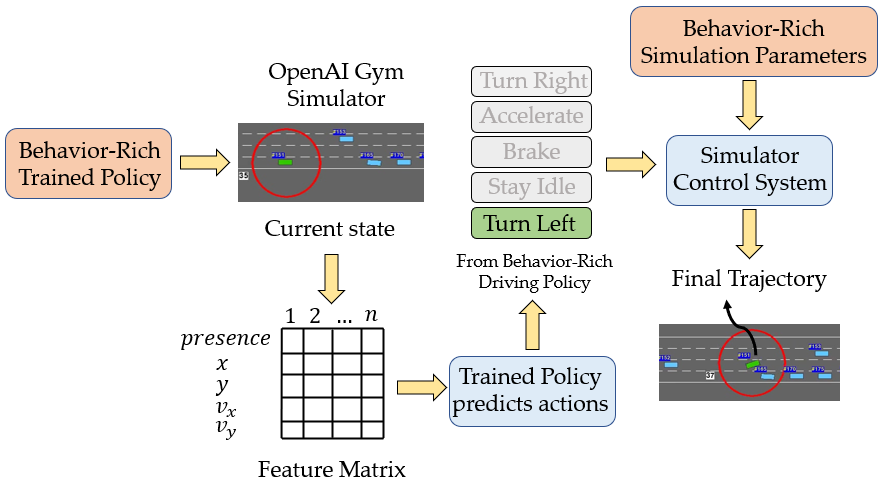}
    \caption{\textbf{Runtime:} We use our behavior-rich trained policy and the final simulation parameters computed using offline training. During an episode at runtime, we use the trained policy to predict the next control of the ego-vehicle given the current state of the environment, represented in the form of a feature matrix. The predicted control (in this case, ``turn left'') is converted into the final local trajectory using the internal controls of the simulator, modified by parameters that consider the behavior of the traffic agents.}
    \label{fig: runtime_figure}
    \vspace{-10pt}
\end{figure}

\subsection{Reinforcement Learning-Based Navigation}
\label{subsec: Reinforcemen_Learning_based_Navigation}
Learning to navigate autonomously in dense traffic environments is a difficult problem due to both the number of agents involved and the uncertainty inherent in their road behavior. The problem becomes even more challenging when the agents demonstrate varying behaviors ranging from conservative behaviors, which can disrupt the smoothness of the traffic flow, to aggressive maneuvers, which increase the probability of an accident. We integrate these behaviors into the simulator using CMetric and thereby generate  more realistic traffic scenarios.

Using these behaviors, we employ a Deep Reinforcement Learning setup to achieve autonomous navigation. Specifically, we use a Multilayer Perceptron (MLP) that receives an observation of the state space as input and implicitly models the behavioral interactions between aggressive and conservative agents and the ego-vehicle. Ultimately, the MLP learns a function that receives a feature matrix, which describes the current state of the traffic as input and returns the optimal $Q$ values of the state space. Finally, the ego-vehicle can use the learned model during evaluation time to navigate its way around the traffic by choosing the best control that corresponds to the maximum $Q$ value for its state at every time step.
\section{Runtime Navigation and Trajectory Computation}
\label{sec: Navigation_at_Runtime}

At runtime, our goal is to perform behaviorally-guided navigation using the RL policy trained offline. The traffic simulator provided by the OpenAI gym environment is modified to account for driver behaviors in a fashion similar to the offline training phase (explained in Section~\ref{sec: Offline_Training}). At each time step, we record the state of the environment, which consists of the positions and velocities of all vehicles in that frame, as described in Section~\ref{subsec: Training A Behavior-Rich Policy}. We store the state in a feature matrix that is passed to the trained behavior-guided policy, computed from the offline training phase. The policy is then used to predict the optimal \blue{ego-control} for the next time step, which is passed to the simulator control system and executed to generate the final trajectory. We present an overview in Figure~\ref{fig: runtime_figure}.

The motion models and controllers used by the simulator compute the final local trajectory, guided by the high-level control predicted by the trained policy. The linear acceleration model is based on the Intelligent Driver Model (IDM)~\cite{treiber2000congested} and is computed via the following kinematic equation,
\begin{equation}
    \dot v_{\alpha} = a\begin{bmatrix}1 - (\frac{v_{\alpha}}{v_0^{\alpha}})^4 - (\frac{s^*(v_{\alpha}, \Delta v_{\alpha})}{s_{\alpha}})^2\end{bmatrix}.
    \label{eq: IDM_acc}
\end{equation}
\noindent Here, the linear acceleration, $\dot v_{\alpha}$, is a function of the velocity $v_{\alpha}$, the net distance gap $s_{\alpha}$, and the velocity difference $\Delta v_{\alpha}$ between the ego-vehicle and the vehicle in front. 
% Equation~\ref{eq: IDM_acc} is a combination of the acceleration on a free road $\dot v_{free} = a[1 - (v/v_0)^{4}]$ (\textit{i.e.} no obstacles) and the braking deceleration, $-a(s^*(v_\alpha,\Delta v_\alpha)/s_\alpha)^2$ (\textit{i.e.} when the ego-vehicle comes in close proximity to the vehicle in front).
The deceleration term depends on the ratio of the desired minimum gap ($s^*(v_\alpha,\Delta v_\alpha)$) and the actual gap ($s_{\alpha}$), where $s^* (v_\alpha,\Delta v_\alpha)= s_0 + vT + \frac{v\Delta v}{2\sqrt{ab}}$. $s_0$ is the minimum distance in congested traffic, $vT$ is the distance while following the leading vehicle at a constant safety time gap $T$, and $a,b$ correspond to the comfortable maximum acceleration and comfortable maximum deceleration, respectively. 
%We use CMetric to learn these parameters for aggressive and conservative driving behaviors (Table~\ref{tab: parameters}).

Complementary to the longitudinal model, the lane changing behavior is based on the MOBIL~\cite{kesting2007general} model. As per this model, there are two key aspects to keep in mind:
\begin{enumerate}
    \item \textit{Safety Criterion}: This condition checks if, after a lane change to a target lane, the ego-vehicle has enough room to accelerate. Formally, we check if the deceleration of the successor $a_\textrm{target}$ in the target lane exceeds a pre-defined safe limit $b_{safe}$:
    \begin{equation*}
        a_\textrm{target} \geq -b_{safe}.
    \end{equation*}
    \item \textit{Incentive Criterion}: This criterion determines the total advantage to the ego-vehicle after the lane change, measured in terms of total acceleration gain or loss. It is computed with the formula
    
    \begin{equation*}
    \tilde{a}_\textrm{ego} - a_\textrm{ego} + p(\tilde{a}_n - a_n + \tilde{a}_o - a_o) > \Delta a_{th},
    \end{equation*}
    where $\tilde{a}_\textrm{ego} - a_\textrm{ego}$ represents the acceleration gain that the ego-vehicle would receive after the lane change. The second term denotes the total acceleration gain/loss of the immediate neighbors (the new follower in the target, $a_n$, and the original follower in the current lane, $a_o$) weighted with the politeness factor $p$. By adjusting $p$, the intents of the drivers can be changed from purely egoistic ($p=0$) to more altruistic ($p=1$). We refer the reader to~\cite{kesting2007general} for further details.
\end{enumerate}

\noindent The lane change is executed if both the safety criterion is satisfied and the total acceleration gain is more than the defined minimum acceleration gain $\Delta a_{th}$. We use CMetric to adjust the parameters $p, \Delta a_{th}, b_{safe}$ in order to introduce aggressive and conservative driving behaviors (Table~\ref{tab: parameters}).

\begin{figure}[t]
\centering
% \resizebox{.95\linewidth}{!}{
\begin{subfigure}[h]{.49\columnwidth}
    \includegraphics[width=\textwidth]{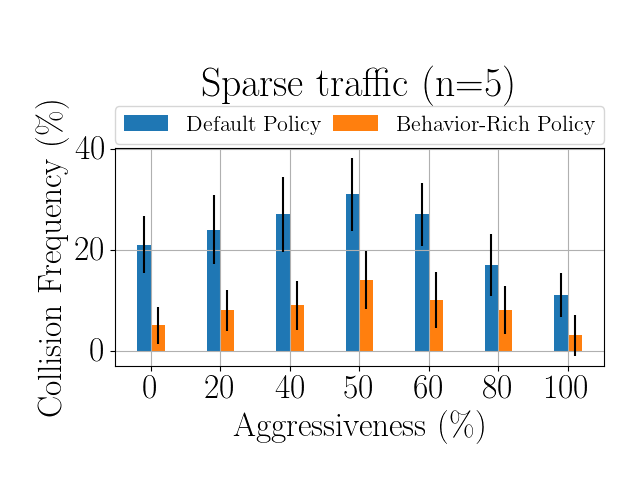}
    \caption{Collision frequency for $n=5$ vehicles.}
    \label{fig: crash_1}
  \end{subfigure}
   \begin{subfigure}[h]{.49\columnwidth}
    \includegraphics[width=\textwidth]{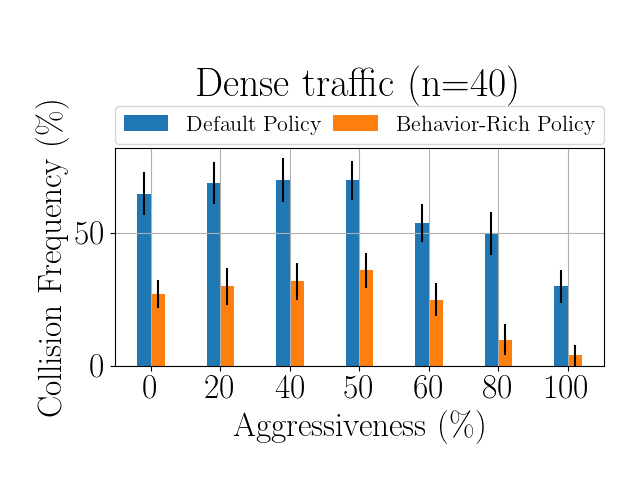}
    \caption{Collision frequency for $n=40$ vehicles.}
    \label{fig: crash_2}
  \end{subfigure}
  \begin{subfigure}[h]{.49\columnwidth}
    \includegraphics[width=\textwidth]{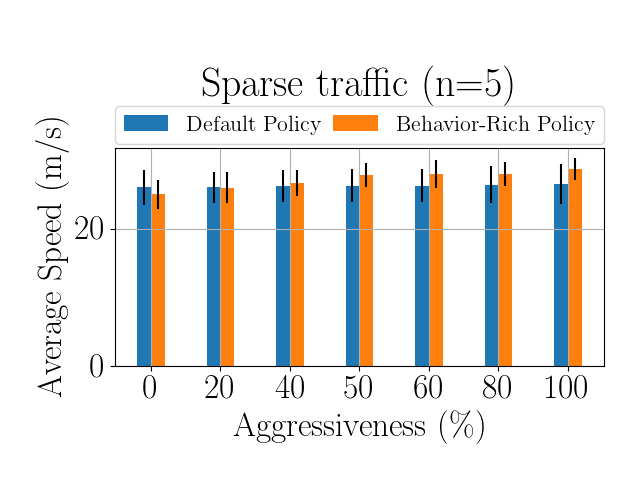}
    \caption{Average speed for $n=5$ vehicles.}
    \label{fig: v_1}
  \end{subfigure}
  \begin{subfigure}[h]{.49\columnwidth}
    \includegraphics[width=\textwidth]{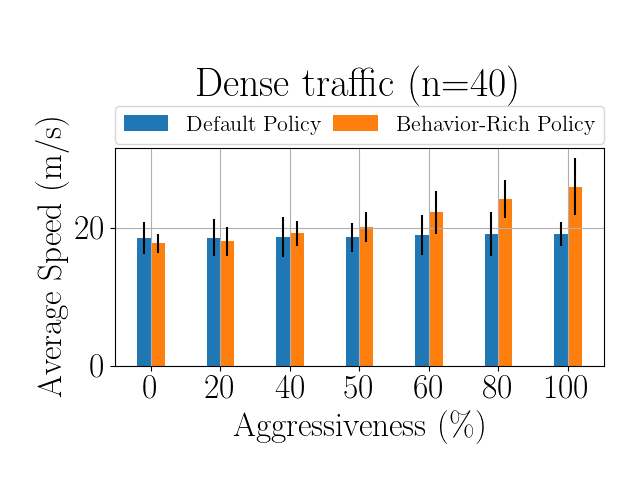}
    \caption{Average speed for $n=40$ vehicles.}
    \label{fig: v_2}
  \end{subfigure}
    \begin{subfigure}[h]{.49\columnwidth}
    \includegraphics[width=\textwidth]{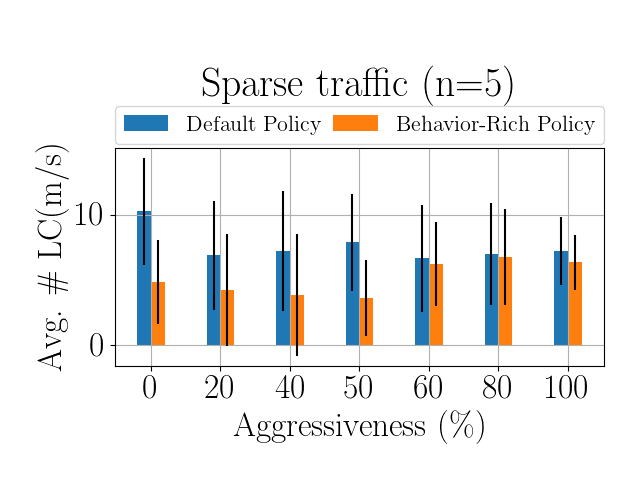}
    \caption{Average number of lane changes for $n=5$ vehicles.}
    \label{fig: lc_1}
  \end{subfigure}
  \begin{subfigure}[h]{.49\columnwidth}
    \includegraphics[width=\textwidth]{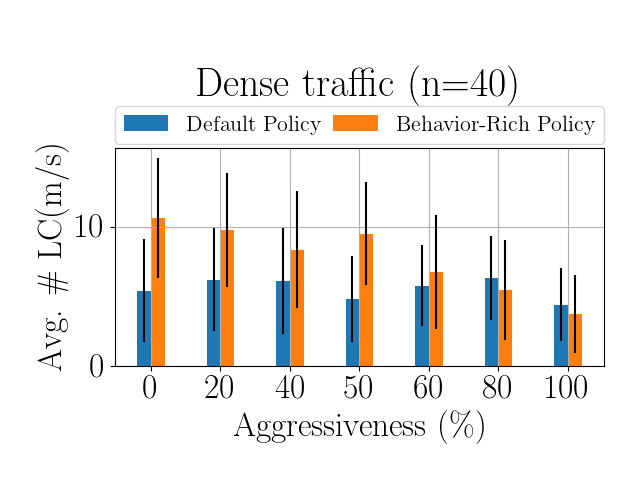}
    \caption{Average number of lane changes for $n=40$ vehicles.}
    \label{fig: lc_2}
  \end{subfigure}
%   }
\caption{\textbf{Evaluation:} The behavior-rich policy is compared with the default policy in traffic scenarios of varying densities and driver behaviors. The behavior-rich policy leads to improved navigation and a reduced number of collisions (Figures~\ref{fig: crash_1},~\ref{fig: crash_2}) by adjusting the speed (Figures~\ref{fig: v_1},~\ref{fig: v_2}) of the ego-vehicle to the behaviors of its neighbors (higher average speed for higher percentages of aggressive neighbors) and its lane changes (Figures~\ref{fig: lc_1},~\ref{fig: lc_2}). When $n=5$ (Figure~\ref{fig: lc_1}), the number of lane changes increases as the ego-vehicle is more confident in performing lane-changing maneuvers due to the sparsity of the traffic. In contrast, the ego-vehicle performs less lane changes in the dense traffic scenario as the percentage of aggressiveness increases, respecting unpredictable surrounding drivers.}
\label{fig: bar_chart}
  \vspace{-10pt}
\end{figure}

% Additionally, the desired velocity $v_0$ is set to $25$ meters per second and $40$ meters per second for the conservative and aggressive vehicle classes, respectively. Finally, the desired velocities for the conservative vehicles were uniformly distributed with a variation of {$\pm$10\%} to increase the heterogeneity in the simulation environment.

\section{Experiments and Results}
\subsection{Simulator}

We used the OpenAI gym-based highway-env simulator~\cite{highway-env} and enriched it with various driving behaviors using CMetric~\cite{cmetric} (as shown in Figure~\ref{fig: offline_training}). Our approach is easily transferable to other state-of-the-art simulators~\cite{SUMO2018,carla}. More specifically, the parameters from the underlying motion models we are using~\cite{treiber2000congested, kesting2007general} are widely used in these simulators. Consequently, CMetric is used to tune these parameters, which can then be used in any of these simulators. In our benchmarks, we consider dense traffic scenarios on a four-lane (lane width = $4m$), one-way highway environment populated by vehicles with various driving behaviors. The ego-vehicle always navigates in the middle of its current lane and can reach speeds ranging from 10 $m/s$ to 30 $m/s$.

% \subsection{Training}

% Our objective is to train a DRL agent to navigate safely and efficiently in dense traffic scenarios populated by vehicles with various driving behaviors. Our method achieves this by predicting the optimal action that the agent needs to take at every time step. For training, w
% We use the Deep Q Learning implementation provided in~\cite{rl-agents} and train a Multilayer Perceptron (MLP) model using PyTorch. To avoid sampling correlated samples during the training of the Deep Q network agent, we use Prioritized Experience Replay~\cite{schaul2016prioritized}. The MLP model receives as input a feature matrix that includes the features for each vehicle present in the current simulation. The simulator provides an observation from the environment in the form of a $F\times V$ matrix, where $V$ is the number of vehicles in the simulation and $F$ is the number of features that constitutes the state space of each vehicle.

% \begin{figure}[t]
%     \centering
%     \includegraphics[width=0.9\columnwidth]{Images/rewards.png}
%     \caption{\textbf{Episode rewards:} This graph shows the training performance of Deep Q Learning with the GCN module. The rewards demonstrate an increasing pattern, as shown by the orange line. This demonstrates that the agent learns to navigate better as it incurs higher rewards based on our reward distribution.}
%     \label{fig: rewards}
%     \vspace{-10pt}
% \end{figure}
\begin{figure*}[t]
\centering
% \resizebox{.95\linewidth}{!}{
\begin{subfigure}[h]{0.32\linewidth}
    \includegraphics[width=\textwidth]{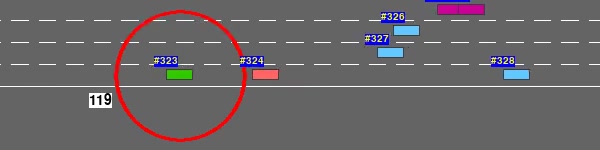}
    \caption{Ego-vehicle approaches aggressive vehicle.}
    \label{fig: slow_down_1}
  \end{subfigure}
   \begin{subfigure}[h]{0.32\linewidth}
    \includegraphics[width=\textwidth]{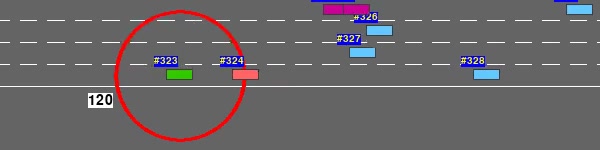}
    \caption{Ego-vehicle slows down.}
    \label{fig: slow_down_2}
  \end{subfigure}
  \begin{subfigure}[h]{0.32\linewidth}
    \includegraphics[width=\textwidth]{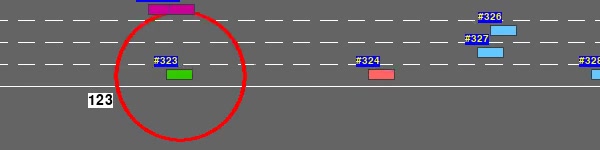}
    \caption{Ego-vehicle slowly accelerates.}
    \label{fig: slow_down_3}
  \end{subfigure}
  \begin{subfigure}[h]{0.32\linewidth}
    \includegraphics[width=\textwidth]{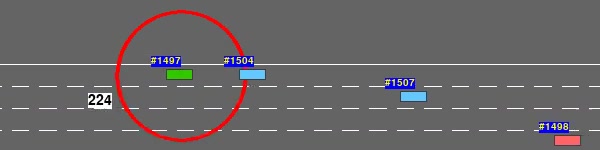}
    \caption{Ego-vehicle approaches conservative vehicle.}
    \label{fig: lane_change_1}
  \end{subfigure}
    \begin{subfigure}[h]{0.32\linewidth}
    \includegraphics[width=\textwidth]{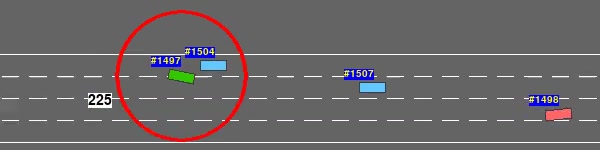}
    \caption{Ego-vehicle switches to adjacent lane.}
    \label{fig: lane_change_2}
  \end{subfigure}
  \begin{subfigure}[h]{0.32\linewidth}
    \includegraphics[width=\textwidth]{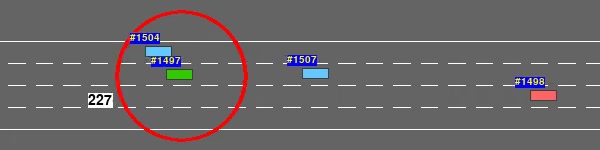}
    \caption{Ego-vehicle overtakes conservative vehicle.}
    \label{fig: lane_change_3}
  \end{subfigure}
%   }
\caption{\textbf{\blue{Behavior-Guided Navigation:}} Interaction with aggressive (\textit{top}) and conservative (\textit{bottom}) vehicles. We indicate the ego-vehicle, aggressive vehicles, and conservative vehicles with green, red, and blue colors, respectively. (\textit{Top}) The ego-vehicle senses that the red vehicle is aggressive and therefore decides to slow down instead of overtaking. (\textit{Bottom}) The ego-vehicle notices the conservative vehicle in front and decides to confidently overtake it. Our approach can automatically generate such trajectories for the ego-vehicle.}
\label{fig: qualitative}
  \vspace{-10pt}
\end{figure*}

\subsection{Implementation and Reward Tuning}

We use the PyTorch-based Deep Q Learning implementation~\cite{rl-agents} to train a $3$-layer MLP model for $3000$ episodes (episode duration=$60$ seconds) in order to tune the reward values using the Epsilon Greedy policy. We initially set $\epsilon = 1$ with $0.05$ decay and train using the Adam optimizer with a learning rate of $\eta=0.0005$ and the Mean Squared Error (MSE) loss function. We save the reward values for every $50$ episodes and test the model for $20$ episodes. To avoid sampling correlated samples during the training of the Deep Q network agent, we use Prioritized Experience Replay~\cite{schaul2016prioritized}. The MLP model receives as input a feature matrix that includes the features for each vehicle present in the current simulation. The simulator provides an observation from the environment in the form of an $F\times V$ matrix, where $V$ is the number of vehicles in the simulation and $F$ is the number of features that constitutes the state space of each vehicle.

% and examine the videos at test time in order to choose the model that best fits our desired behavior. 
Our goal is to achieve a low collision frequency while simultaneously maintaining the speed of the ego-vehicle at levels that guarantee efficient navigation and smooth traffic flow. We manually tune the reward values monitoring the collision frequency and the average speed of the ego-vehicle at test time. The same reward values are used for all types of traffic considered. We increase the reward value $r_{HS}$ when the ego-vehicle is moving too slowly and decrease the reward for collisions $r_{C}$, thereby penalizing collisions, when the ego-vehicle is prioritizing higher speed over collision avoidance. We also keep the value of $r_{RL}$ at a low level so that the ego-vehicle does not stay too long in the right lane. Furthermore, we tune $r_{LC}$ based on the pattern, according to which the ego-vehicle performs a lane change. More specifically, we visually inspect the performance at test time and increase $r_{LC}$ when the ego-vehicle tends to stay too long behind other vehicles while the adjacent lanes are empty to encourage overtaking maneuvers. In contrast, we decrease $r_{LC}$ when the ego-vehicle performs many "unnecessary" lane changes, since they disrupt the smoothness of the generated trajectory and increase the possibility of an accident.
% Fig.~\ref{fig: rewards} shows the total reward per episode which follows an increasing trend guided by the orange line.
% If we observe too many unnecessary lane changes, we decrease $r_{LC}$, and if we do not observe any lane change at all when approaching another vehicle, we increase it.

% We experiment with various network structures and decided on a $3$-layer Graph Convolutional network with ReLu activations and a fully connected final layer with dropout that predicts the computed $Q$ values.
% For balancing exploration with exploitation we employ the Epsilon Greedy policy, according to which at every time step the agent explores by choosing a random action with probability $\epsilon$, and exploits by choosing the maximum $Q$ value that corresponds to the next state with probability $1-\epsilon$. 
% The value of $\epsilon$ initially is $1$, where actions get chosen entirely in random and gradually decays to $0.05$, where the agent almost certainly executes the best possible action according to the Deep Q Learning algorithm.
% Finally, we use the Adam optimizer with a learning rate of $\eta=0.0005$ and the Mean Squared Error (MSE) loss function for the $Q$ values that the network outputs. Fig.~\ref{fig: rewards} shows the total reward per episode which follows an increasing trend guided by the orange line despite the reward values themselves being fuzzy.

\subsection{Evaluation and Results}
We evaluate the performance of our method in both sparse and dense traffic scenarios by varying the number of aggressive and conservative agents. We apply three metrics for evaluation averaged over multiple $100$-episode runs:

\begin{itemize}
    \item \textit{Collision frequency} (\%) of the ego-vehicle measured as a percentage over the total test runs.
    \item \textit{Average speed} ($m/s$) of the ego-vehicle, as it captures distance per second covered in a varying time interval.
    \item \textit{Number of lane changes} performed by the ego-vehicle on average during the given duration.
    % We notice that our approach based on GCN results in an approximately $50$\% reduction in the number of lane changes compared to MLP.
\end{itemize} 

\noindent In terms of density, we consider a sparse traffic scenario ($n=5$) and a dense traffic scenario ($n=40$). \blue{In terms of driver behaviors, we examine seven distinct types of traffic consisting of \{$0\%$, $20\%$, $40\%$, $50\%$, $60\%$, $80\%$, $100\%$\} aggressive drivers. The remaining drivers in each scenario are conservative agents. For example, in the second scenario, we consider $20\%$ aggressive agents and $80\%$ conservative agents.} Figure~\ref{fig: bar_chart} shows the performance of our behavior-rich policy compared to the default policy generated by the simulator, which does not consider driver behavior. The default policy was trained using an identical training routine in a simulator environment where all the cars follow a default built-in behavior based on the IDM model~\mbox{\cite{treiber2000congested}}. In these experiments, the vehicles are observable if they are within $d=180m$ from the ego-vehicle.
%\hl{When the ego-vehicle moves with a maximum speed of $30m/s$, this distance ($180m$) can be interpreted as an ability to look 6 seconds ahead ($30m/s\cdot 6s=180m/s$).

In both the sparse (Figure~\ref{fig: crash_1}) and the dense (Figure~\ref{fig: crash_2}) scenario, our behavior-rich policy leads to a significantly lower collision rate than the baseline policy of the simulator. We observe fewer collisions when all the road-agents are either aggressive \blue{($100\%$ Aggressiveness)} or conservative \blue{($0\%$ Aggressiveness)} than when the traffic is mixed \blue{($20\%,40\%,60\%,80\%$ Aggressiveness)}. This occurs because the ego-vehicle gets confused by the combination of slow-moving vehicles and aggressive agents. To further improve performance, we would need a different set of rewards to be tuned for each specific type of traffic examined. \blue{The default policy has a lower collision frequency when the behaviors are aggressive, both in the sparse scenario (Figure~\ref{fig: crash_1}) and in the dense scenario (Figure~\ref{fig: crash_2}). This occurs because the ego-vehicle, using the default policy, drives with reduced speed, compared to the aggressive vehicles that overspeed and drive ahead. This creates an empty space around the ego-vehicle as it is “left behind” and reduces the collision frequency compared to the mixed and conservative traffic scenarios when there are vehicles around the ego-vehicle driving at comparable speed.}

The average speed of the ego-vehicle increases (Figures~\ref{fig: v_1},~\ref{fig: v_2}) as the traffic ranges from conservative to aggressive with our behavior-rich policy, as slow-moving conservative vehicles tend to slow the ego-vehicle down. In contrast, the speed in the default policy remains almost unchanged, indicating that the ego-vehicle is unable to adjust to the behaviors of the road-agents and maintains a uniform behavior. Finally, the number of lane changes performed by the ego-vehicle increases as the traffic becomes more aggressive for the sparse traffic scenario (Figure~\ref{fig: lc_1}) and the behavior-rich policy. In that case, the ego-vehicle leverages the sparsity of the traffic and, influenced by its aggressive neighbors, attempts lane-changing maneuvers more confidently. In  the dense traffic scenario (Figure~\ref{fig: lc_2}), the number of lane changes of the ego-vehicle decreases as the surrounding agents become more aggressive. We show qualitative results in Figure~\ref{fig: qualitative}.

\begin{figure}[t]
\centering
\begin{subfigure}[h]{0.49\columnwidth}
    \includegraphics[width=\textwidth]{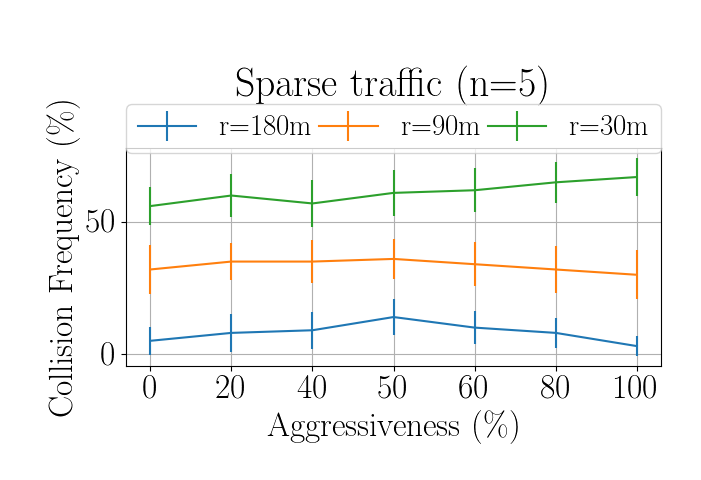}
    \caption{\blue{Collision frequency for different ego-perception abilities in sparse traffic.}}
    \label{fig: partial_sparse}
  \end{subfigure}
   \begin{subfigure}[h]{0.49\columnwidth}
    \includegraphics[width=\textwidth]{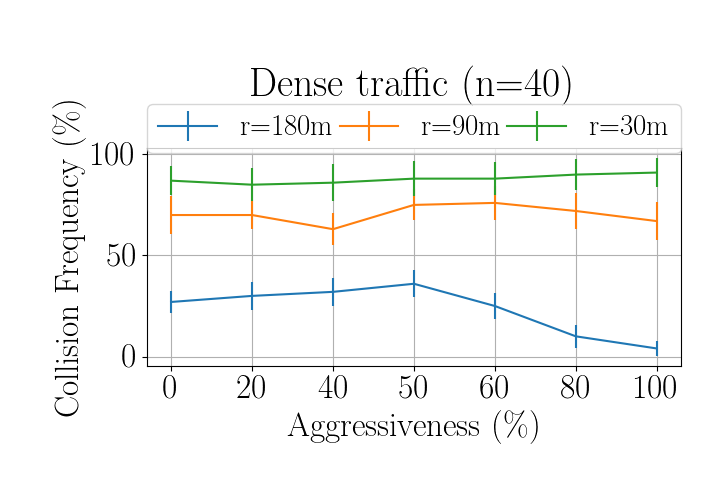}
    \caption{\blue{Collision frequency for different ego-perception abilities in dense traffic.}}
    \label{fig: partial_dense}
  \end{subfigure}
%   }
\caption{\textbf{\blue{Partial Observability:}} \blue{We model partial observability by reducing the perception ability of the ego-vehicle, adding vehicles to the state space only when they are within a specified radius from the ego-vehicle. We present results in terms of collisions from experiments that follow a behavior-rich policy and consider a perception radius around $180m$, $90m$, and $30m$ from the ego-vehicle. The navigation performance drops as the radius decreases and the number of crashes is larger when the ego-vehicle has to navigate around a larger number of agents (Figure~\ref{fig: partial_dense}).}}
\label{fig: partial_observability}
  \vspace{-10pt}
\end{figure}

In addition to the initial experiments that consider a perception radius of $d_1=180m$ around the ego-vehicle, we present results (Figure~\mbox{\ref{fig: partial_observability}}) using our behavior-rich simulation technique for limited sensing abilities corresponding to radii of $d_2=90m$ and $d_3=30m$, respectively. The results show that the navigation performance drops as the perception radius decreases, especially for the dense traffic scenario~\ref{fig: partial_dense}, where the ego-vehicle has to deal with a larger number of agents. In summary, our \blue{new simulation technique combined with this DRL-based navigation scheme} offers the following benefits:

\begin{itemize}[noitemsep]
\item {\em Increased Safety}: Our behavior-rich policy results in a significantly lower collision frequency than the default policy baseline.

\item {\em Behavior-Aware Trajectories}: Our behavior-rich policy generates behavior-aware trajectories. The ego-vehicle learns to adjust its behavior based on the behavior of its neighbors, both in terms of speed regulation and the number of executed lane changes. 
% In practice, the ego-vehicle increases its speed as the percentage of aggressive road-agents increases to be able to compete with them. At the same time it tends to decrease its lane changes to respect unpredictable drivers, but confidently performs overtaking maneuvers in sparse traffic scenarios.

\end{itemize}

\label{subsec: comparison_with_prior_work}
\section{Conclusion, Limitations, and Future Work}

We proposed a new simulation technique for enriching traffic simulators with behavior-rich trajectories corresponding to varying levels of aggressiveness. Based on this technique, we presented a navigation scheme that is based on deep reinforcement learning and considers driver behavior to perform \textit{behaviorally-guided} navigation. Our preliminary results in many benchmarks are encouraging and our approach can handle scenarios with high traffic density and aggressive behaviors. Our behavior generation module can be easily transferred to any state-of-the-art parametric traffic simulator and can generate synthetic traffic datasets consisting of realistic behavior-rich trajectories, which can provide fruitful and diverse training scenarios to state-of-the-art data-driven approaches.
% We use a principled driver behavior model, CMetric, that integrate concepts from the fields of graph theory, machine learning, and traffic psychology to model the interactions between the autonomous vehicle and the human drivers. The output of our approach is a trained policy for autonomous vehicles in environments in which human drivers perform aggressive maneuvers such as overtaking, overspeeding, weaving, and sudden lane-changes. We tested our approach extensively using a recently developed simulator for dense urban traffic.

Our method has several limitations. For example, in some cases, the ego-vehicle avoids aggressive vehicles by decelerating and performing fewer lane changes, thereby acting conservatively. In some scenarios, it may be necessary to execute non-conservative actions such as increasing the frequency of lane changes or following local traffic norms. \blue{Furthermore, the performance of our approach drops as the perception radius of the ego-vehicle decreases, which means that it would require very good sensing for it to be applicable to real-world scenarios.} In terms of future work, we plan to extend our approach to different environments like roundabouts, intersections, and parking lots and perform more detailed evaluation. Furthermore, we plan to develop efficient navigation techniques to handle heterogeneous traffic-agents such as pedestrians or bicycles.
% Finally, we would like evaluate our performance in scenarios corresponding to complex urban environments and improve the navigation performance

% \section*{Acknowledgements}

%This work was supported in part by ARO Grants W911NF1910069, W911NF2110026, and U.S. Army Grant No. W911NF2120076.

%%%%%%%%%%%%%%%%%%%%%%%%%%%%%%%%%%%%%%%%%%%%%%%%%%%%%%%%%%%%%%%%%%%%%%%%%%%%%%%%

\bibliographystyle{IEEEtran}
\bibliography{refs}

\clearpage
\begin{appendices}
\section{CMetric: Mapping Trajectories to Behavior}
\label{appendix: CMetric}

\subsection{Centrality Measures}
\label{sec: centrality}
% \subsection{Use of Centrality Measures - What are Centrality Measures?}

In graph theory and network analysis, centrality measures~\cite{rodrigues2019network} are real-valued functions $\zeta:\mathcal{V}\longrightarrow \mathbb{R}$, where $\mathcal{V}$ denotes the set of vertices and $\mathbb{R}$ denotes a scalar real number that identifies key vertices within a graph network. So far, centrality functions have been restricted to identifying influential personalities in online social media networks~\cite{cen-socialmedia} and key infrastructure nodes in the Internet~\cite{cen-internet}, to rank web-pages in search engines~\cite{cen-pagerank}, and to discover the origin of epidemics~\cite{cen-epidemic}. There are several types of centrality functions. The ones that are of particular importance to us are the degree centrality and the closeness centrality denoted as $\pd$ and $\pc$, respectively. These centrality measures are defined in~\cite{cmetric} (See section III-C). Each function measures a different property of a vertex. Typically, the choice of selecting a centrality function depends on the current application at hand. In this work, the closeness centrality and the degree centrality functions measure the likelihood and intensity of specific driving styles such as overspeeding, overtaking, sudden lane-changes, and weaving~\cite{cmetric}.

% we show that centrality functions can measure the likelihood and intensity of different driver styles such as overspeeding, overtaking, sudden lane-changes, and weaving.
%  (See Table~\ref{tab: centrality_relationship}). 
 
% There are several types of centrality functions. The ones that are of particular importance to us are the degree centrality and the closeness centrality denoted as $\pd$ and $\pc$, respectively. These centrality measures are defined in~\cite{cmetric} (See section III-C). Each function measures a different property of a vertex. Typically, the choice of selecting a centrality function depends on the current application at hand. For instance, in the examples mentioned earlier, the eigenvector centrality is used in search engines while the degree centrality is appropriate for measuring the popularity of an individual in social networks. Likewise, the centrality and the degree different centrality functions measure the likelihood and intensity of specific driving styles such as overspeeding, overtaking, sudden lane-changes, and weaving~\cite{cmetric}. 

\subsection{Algorithm}
\label{sec: approach}

Here, we present the main algorithm, called \textit{CMetric}. CMetric maps vehicle trajectories to specific styles by computing the likelihood and intensity of the latter using the definitions of the centrality functions. The specific styles are then used to assign global behaviors~\cite{sagberg2015review}. We summarize the CMetric algorithm as follows:

% \textbf{CMetric Algorithm:}

\begin{enumerate}
    \item Obtain the positions of all vehicles using localization sensors deployed on the autonomous vehicle and form traffic-graphs at each time-step.
    
    % \item At each time instance, we compute the Laplacian matrix $L_t$ using Equation~\ref{eq: Lt}.
    
    \item Compute the closeness and degree centrality function values for each vehicle at every time-step.
    
    \item Perform polynomial regression to generate uni-variate polynomials of the centralities as a function of time. 
    
    \item Measure likelihood and intensity of a specific style for each vehicle by analyzing the first- and second-order derivatives of their centrality polynomials.
    
\end{enumerate}

\noindent We begin by forming the traffic-graphs for each frame and use the definitions in~\cite{cmetric} to compute the discrete-valued centrality measures. Since centrality measures are discrete functions, we perform polynomial regression using regularized Ordinary Least Squares (OLS) solvers to transform the two centrality functions into continuous polynomials, $\pc$ and $\pd$, as a function of time. We describe polynomial regression in detail in the following subsections. We compute the likelihood and intensity of specific styles by analyzing the first- and second-order derivatives of $\pc$ and $\pd$.

% \subsection{Traffic-Graph Formation Using Localization Sensors}
% \label{subsec: step_1}
% Our algorithm assumes that global coordinates of all vehicles are provided, however, state-of-the-art results in localization systems show that the error can be reduced to the order of millimeters. Given the input coordinates, we assign a coordinate through a $2$-dimensional tuple $[x,y]^\top$ to a vertex in the traffic-graph. At any time-step, once the assignment for all the vehicles in that frame is completed, we proceed to generating the adjacency and laplacian matrices of the resulting graph, as described in Section~\ref{subsec: DGG}.

% \subsection{Computing Centrality Values}
% \label{subsec: step_1}
% Our algorithm assumes that global coordinates of all vehicles are provided, however, state-of-the-art results in localization systems show that the error can be reduced to the order of millimeters. Given the input coordinates, we assign a coordinate through a $2$-dimensional tuple $[x,y]^\top$ to a vertex in the traffic-graph. At any time-step, once the assignment for all the vehicles in that frame is completed, we proceed to generating the adjacency and laplacian matrices of the resulting graph, as described in Section~\ref{subsec: DGG}.

\subsection{Polynomial Regression}
\label{subsec: polynomial_regression}
 
In order to study the behavior of the centrality functions with respect to how they change with time, we convert the discrete-valued $\zeta[t]$ into continuous-valued polynomials $\zeta(t)$, using which we calculate the first- and second-order derivatives of the centrality functions as explained in Section~\ref{subsec: SLE_SIE}.

In this work, we choose a quadratic\footnote{A polynomial with degree $2$.} centrality polynomial can be expressed as $\zeta(t) = \beta_0 + \beta_1 t + \beta_2t^2$, as a function of time. Here, $\beta = [\beta_0 \ \beta_1 \ \beta_2]^\top$ are the polynomial coefficients. These coefficients can be computed using ordinary least squares (OLS) equation as follows,

% \begin{equation}
%     \beta = \argmin_\beta \Vts{\zeta - M\beta}
% \label{eq: optimization_problem}
% \end{equation}
% {\small
\begin{equation}
    % \begin{split}
        % M \beta &= \zeta^i  \\
        % (M^\top M) \beta &= M^\top \zeta^i  \\
        \beta = \inv{(M^\top M)}M^\top \zeta^i
    % \end{split}
    \label{eq: noiseless_OLS}
\end{equation}

\noindent Here, $M \in \mathbb{R}^{T\times(d+1)}$ is the Vandermonde matrix~\cite{manocha1992algebraic}.
and is given by, 

\[
M = \begin{bmatrix}
    1 & t_1 & t^2_1 & \dots  & t^d_1 \\
    1 & t_2 & t^2_2 & \dots  & t^d_2 \\
    \vdots & \vdots & \vdots & \ddots & \vdots \\
    1 & t_T & t^2_T & \dots  & t^d_T
\end{bmatrix}
\]

% \noindent where $d = 2 \ll T$ is the degree of the resulting centrality polynomial~\cite{manocha1992algebraic}.
% The resulting estimator for $\beta$ is given by $\beta = \inv{(M^\top M)}M^\top \zeta^i$.

% \begin{figure}[t]
%     \centering
%     \includegraphics[width = \columnwidth]{img/Condition_Number.png}
%     \caption{\textbf{Robustness to Noise:} We show that by regularizing the noisy OLS system given by Equation~\ref{eq: noisy_OLS}, we can reduce the original condition number (red curve) while at the same time upper bounding the reduced condition number (blue curve) by $\delta \longrightarrow 0$. The reduced condition number helps stabilize the noisy estimator $\tilde \beta$. }
%     \label{fig: condition_number}
% \end{figure}

\subsection{Style Likelihood and Intensity Estimates}
\label{subsec: SLE_SIE}
In the previous sections, we used polynomial regression on the centrality functions to compute centrality polynomials. In this section, we analyze and discuss the first and second derivatives of the degree centrality, $\pd$, and closeness centrality, $\pc$, polynomials. Based on this analysis, which may vary for each specific style, we compute the Style Likelihood Estimate (SLE) and Style Intensity Estimate (SIE)~\cite{cmetric}, which are used to measure the probability and the intensity of a specific style.

\paragraph{Overtaking/Sudden Lane-Changes}
Overtaking is when one vehicle drives past another vehicle in the same or an adjacent lane, but in the same direction. The closeness centrality increases as the vehicle navigates towards the center and vice-versa. The SLE of overtaking can be computed by measuring the first derivative of the closeness centrality polynomial using $\textrm{SLE}(t) = \abs*{\frac{\partial \pc}{\partial t}}$. The maximum likelihood $\textrm{SLE}_\textrm{max}$ can be computed as $\textrm{SLE}_{\textrm{max}} = \max_{t \in \Delta t}{\textrm{SLE}}(t)$. The SIE of overtaking is computed by simply measuring the second derivative of the closeness centrality using~$ \textrm{SIE}(t) = \abs*{\frac{\partial^2 \pc}{\partial t^2}}$. Sudden lane-changes follow a similar maneuver to overtaking and therefore can be modeled using the same equations used to model overtaking. 
% We visualize the use of closeness centrality to model overtaking in Figure~\ref{fig: closeness_demo}.

\paragraph{Overspeeding}

The degree centrality can be used to model overspeeding. As $A_t$ is formed by adding rows and columns to $A_{t-1}$, the degree of the $i^\textrm{th}$ vehicle (denoted as $\theta_i$) is calculated by simply counting the number of non-zero entries in the $i^\textrm{th}$ row of $A_t$. Intuitively, a drivers that are overspeeding will observe new neighbors along the way (increasing degree) at a higher rate than conservative, or even neutral, drivers. Let the rate of increase of $\theta_i$ be denoted as $\theta_i^{'}$. By definition of the degree centrality and construction of $A_t$, the degree centrality for an aggressively overspeeding vehicle will monotonically increase. Conversely, the degree centrality for a conservative vehicle driving at a uniform speed or braking often at unconventional spots such as green light intersections will be relatively flat. 
% Figure~\ref{fig: degree_demo} visualizes how the degree centrality can distinguish between an overspeeding vehicle and a vehicle driving at a uniform speed.  
Therefore, the likelihood of overspeeding can be measured by computing,
\[\textrm{SLE}(t) = \abs*{\frac{\partial \pd}{\partial t}}\]

\noindent Similar to overtaking, the maximum likelihood estimate is given by $\textrm{SLE}_{\textrm{max}} = \max_{t  \in \Delta t}{\textrm{SLE}}(t)$.

\paragraph{Weaving} A vehicle is said to be weaving when it ``zig-zags'' through traffic. Weaving is characterized by oscillation in the closeness centrality values between low values towards the sides of the road and high values in the center. Mathematically, weaving is more likely to occur near the critical points (points at which the function has a local minimum or maximum) of the closeness centrality polynomial. The critical points $t_c$ belong to the set $\mathcal{T} =\big\{ t_c \big | \frac{\partial \zeta_c(t_c)}{\partial t} = 0 \big\}$. Note that $\mathcal{T}$ also includes time-instances corresponding to the domain of constant functions that characterize conservative behavior. We disregard these points by restricting the set membership of $\mathcal{T}$ to only include those points $t_c$ whose $\varepsilon-$sharpness~\cite{dinh2017sharp} of the closeness centrality is non-zero. The set $\mathcal{T}$ is reformulated as follows,
% {\small
\begin{equation}
    % \zeta_d(u) = \vts{ \{ A(u,v)\in A_{u,:} \land  \nu_u \geq \nu_v \} },
    \begin{aligned}
    \mathcal{T} = \Bigg \{ t_c \bigg | \frac{\partial \zeta_c(t_c)}{\partial t} = 0 \Bigg \} \\
    \textrm{s.t.} \max_{t \in \mathcal{B}_\varepsilon(t_c)} \frac{\partial \pc}{\partial t}  \neq \frac{\partial \zeta_c(t_c)}{\partial t}\\
    \end{aligned}
    % \zeta_d(u) = 
    \label{eq: weaving}
\end{equation}
% }
% \[\max_{t \in \mathcal{B}_\varepsilon(t^{*})} \textrm{SLE}(t) - \textrm{SLE}(t^{*}) \neq 0,\]
\noindent where $\mathcal{B}_\varepsilon (y) \in \mathbb{R}^d$ is the unit ball centered around a point $y$ with radius $\varepsilon$. The SLE of a weaving vehicle is represented by $\vts{\mathcal{T}}$, which represents the number of elements in $\mathcal{T}$. The $\textrm{SIE}(t)$ is computed by measuring the $\varepsilon-$sharpness value of each $t_c \in \mathcal{T}$. 
% \begin{figure}[t]
% \centering
% \begin{subfigure}[h]{\linewidth}
%   \includegraphics[width=\linewidth]{img/closeness_demo_new.png}
% \caption{\textbf{Sudden Lane Change/Weaving: }The closeness centrality for the green agent \#0 decreases away from the center (scenario A) and increases towards the center (scenario B).}
% \label{fig: closeness_demo}
% \end{subfigure}
% %
% \begin{subfigure}[h]{\linewidth}
%   \includegraphics[width=\linewidth]{img/decision_making_demo.png}
% \caption{\textbf{Overspeeding: }The degree centrality monotonically increases. The rate of increase indicates the scale of aggressiveness, with higher rates of change indicating more aggressively overspeeding drivers. The blue plot corresponds to agent \#0, while the red plot corresponds to agent \#10. Note that agent \#10 is being overtaken by agents that appeared earlier (\#0 and \#8) due to it's conservative speed. Consequently, it's degree centrality plot is constant at $0$.}
% \label{fig: degree_demo}
% \end{subfigure}
% % \centering
% \caption{Simulation results of various driving styles (overspeeding, weaving, lane changing) with their corresponding centrality values. The red circle represents the radius for the agent \#0.}
% \label{fig: simulator_figs}
% % \vspace{-10pt}
% \end{figure}
\paragraph{Conservative Vehicles}

Conservative vehicles, on the other hand, are not inclined towards aggressive maneuvers such as sudden lane-changes, overspeeding, or weaving. Rather, they tend to stick to a single lane~\cite{ahmed1999modeling} as much as possible, and drive at a uniform speed~\cite{sagberg2015review} below the speed limit. Correspondingly, the values of the closeness and degree centrality functions in the case of conservative vehicles remain constant. Mathematically, the first derivative of constant polynomials is $0$. The SLE of conservative behavior is therefore observed to be approximately equal to $0$. Additionally, the likelihood that a vehicle drives uniformly in a single lane during time-period $\Delta t$ is higher when,

\[\abs*{\frac{\partial \pc}{\partial t}} \approx 0 \  \textrm{and} \ \max_{t \in \mathcal{B}_\varepsilon(t^{*})} \textrm{SLE}(t) \approx \textrm{SLE}(t_c) .\]

\noindent The intensity of such maneuvers will be low and is reflected in the lower values for the SIE. 

\begin{table}[t]
\caption{\textbf{Comparison with Prior Work:} We compare B-GAP with prior DRL-based navigation methods. For fair comparison, we compare \% collisions along with the number of vehicles and the number of episodes used for training.}
\centering
\resizebox{\columnwidth}{!}{%
\begin{tabular}{lccc} 
\toprule
Methods & \# vehicles & \# episodes & \% collisions   \\
\midrule
DDPG~\cite{highway1} & $4$ & $4000$ & $10.13$ \\
PS-DDPG~\cite{highway2} & $5$ & $600$ & $11.40$\\
% Kaushik et al. 2018c~\cite{highway3} & $4$ & $10$ \\
% Liu et al.~\cite{inter1} &Intersection & $15$ &$4000$& $6.25$ \\
% Leurent et al.~\cite{inter2} &Intersection& - & $3000$ & $29.90$\\
\textbf{B-GAP } & $5$ & $3000$ & $\textbf{3.00}$ \\
\bottomrule
\end{tabular}
}
\label{tab: comparison}
\vspace{-10pt}
\end{table}
\subsection{Comparison with DRL-Based Navigation Methods}
We compare our navigation scheme with state-of-the-art DRL-based methods that train navigation policies in highway environments consisting of $4-5$ vehicles, with a maximum density of $5$ vehicles. However, these methods do not consider aggressive agents and have not been tested in complex scenarios with high numbers of vehicles and varying behaviors, as we have evaluated (with $n=40$) in Figure~\ref{fig: bar_chart}. 
%Neither approach presents a direct comparison to our method. However,
%A key aspect of our approach is that unlike prior work, our navigation policy is designed for dense traffic containing aggressive agents. 
% Furthermore, many prior works~\cite{highwaync1,highwaync2,highwaync4} assume a controller or an oracle to guarantee collision-free trajectories. Our approach makes no such assumptions. Instead, o
Our navigation method results in very few collisions in high density scenarios ($n=40$ as compared to maximum $5$ used in~\cite{highway2}) with aggressive agents (Figure~\ref{fig: bar_chart}).

In Table~\ref{tab: comparison}, we compare our approach with these methods in terms of \% collisions. For fair comparison, we use $n=5$ in the aggressive environment since prior methods~\cite{highway1, highway2} have also tested with $4-5$ vehicles, respectively. We also include the number of vehicles and training episodes since the performance of these methods depends on them. \blue{It should be noted that these approaches use a continuous action space, while we use a discrete action space.}
\end{appendices}

\end{document}